\documentclass{article}
\usepackage{arxiv}

\usepackage{hyperref}       % hyperlinks
\usepackage{url}            % simple URL typesetting
\usepackage{amsfonts}       % blackboard math symbols
\usepackage{graphicx}
\usepackage{bm}
\usepackage{algorithm}
\usepackage{algpseudocode}
\usepackage{amsmath}
\usepackage{makecell}
\usepackage{support-caption}
\usepackage{subcaption}
\usepackage{multirow}
\usepackage{natbib}

\title{Flexible, Non-parametric Modeling Using Regularized Neural Networks}
\author{
  Oskar Allerbo \\
  Mathematical Sciences\\
  University of Gothenburg and Chalmers University of Technology\\
  \texttt{allerbo@chalmers.se} \\
  ~\\
  Rebecka J\"ornsten\\
  Mathematical Sciences\\
  University of Gothenburg and Chalmers University of Technology\\
  \texttt{jornsten@chalmers.se} 
}
\begin{document}
\maketitle

\begin{abstract}
Non-parametric, additive models are able to capture complex data dependencies in a flexible, yet interpretable way. However, choosing the format of the additive components often requires non-trivial data exploration. Here, as an alternative, we propose PrAda-net, a one-hidden-layer neural network, trained with proximal gradient descent and adaptive lasso. 
PrAda-net automatically adjusts the size and architecture of the neural network to reflect the complexity and structure of the data. 
The compact network obtained by PrAda-net can be translated to additive model components, making it suitable for non-parametric statistical modelling with automatic model selection. We demonstrate PrAda-net on simulated data, where we compare the test error performance, variable importance and variable subset identification properties of PrAda-net to other lasso-based regularization approaches for neural networks. We also apply PrAda-net to the massive U.K.\ black smoke data set, to demonstrate how PrAda-net can be used to model complex and heterogeneous data with spatial and temporal components. 
In contrast to classical, statistical non-parametric approaches, PrAda-net requires no 
preliminary modeling to select the functional forms of the additive components, yet still results in an interpretable model representation.
\end{abstract}

\textbf{Keywords:} additive models, model selection, non-parametric regression, neural networks, regularization, adaptive lasso

\section{Introduction}
Non-parametric, additive models combine the flexibility of non-parametric regression (e.g.\ splines and smoothers) with the interpretability of an additive model structure \citep{hastie1990generalized}. 
For non-parametric functions $\{f_j\}_{j=1}^r$, the model can be formulated as
$$ \mathbb{E}[y]=\sum_{j=1}^r f_j(x_{j}).  $$
For more flexible integration of explanatory variables, \citet{friedman1981projection} proposed projection pursuit regression (PPR), where the non-parametric additive components are formed from a component-specific projection of explanatories: 
\begin{equation}
\label{eq:PPR}
\mathbb{E}[y]=\sum_{j=1}^r f_j(\bm{\beta_j}^T \bm{x}).
\end{equation}
By allowing for the number of components, $r$, in Equation \ref{eq:PPR} to grow, more and more complex data structures can be captured by the model.

PPR is very flexible, but loses some of the interpretability of the additive model as the projection parameters $\bm{\beta_j}$ may combine several covariates into additive components in a way that is often difficult to untangle. While a sparse projection could alleviate this fact, it is not trivial to combine with the selection and flexibility of the non-parametric functions, $f_j$.
In practice, especially for high-dimensional and complex data,  it is therefore common to use data exploration, marginal or partial modeling and domain knowledge to pre-define the subsets of covariates (e.g.\ potential interactions, component content) to combine into additive components, as well as to pre-select the family of functional components (e.g.\ which type of spline).

\begin{figure}
  \centering
  \includegraphics[width=0.4\textwidth]{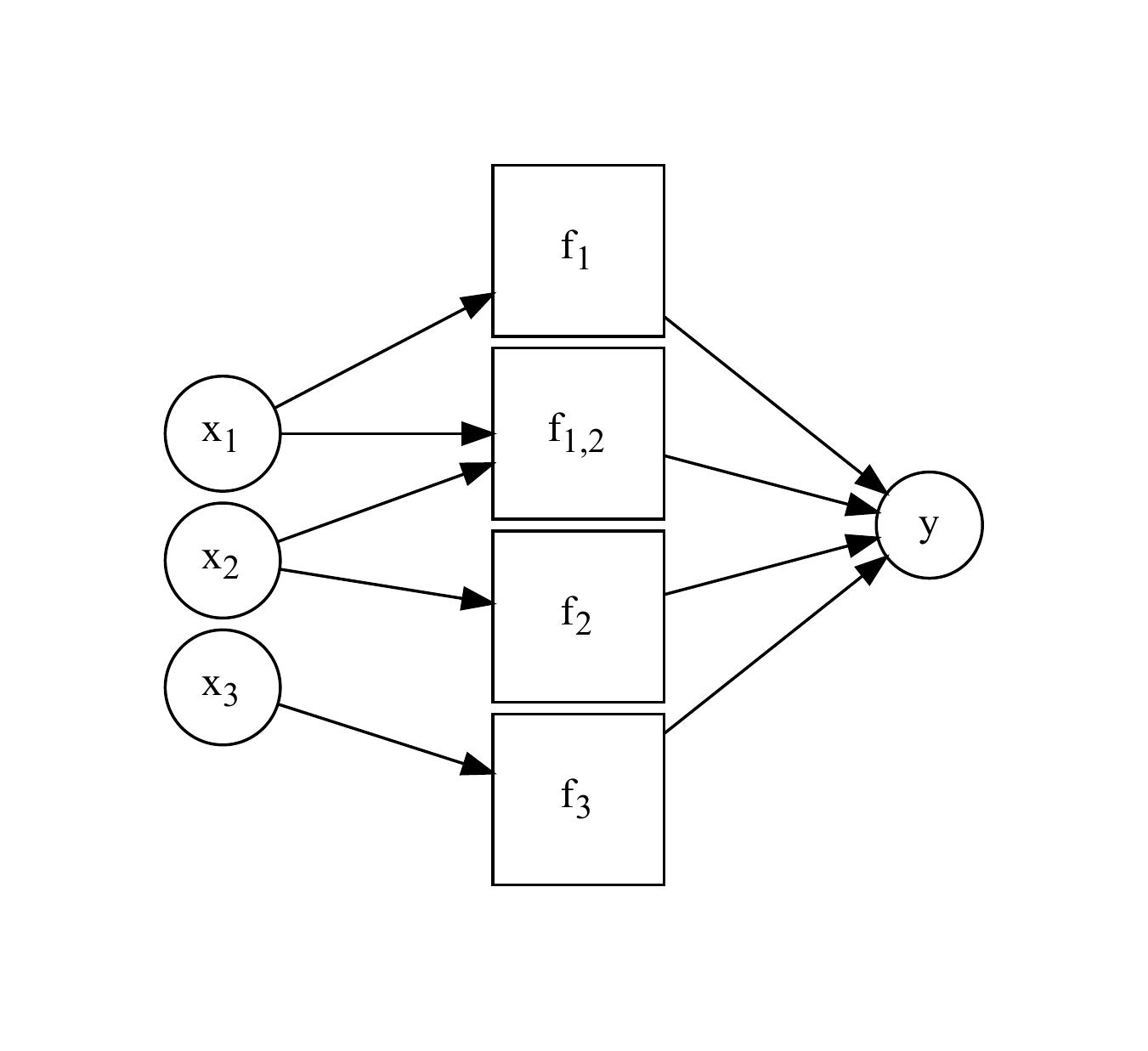}
  \caption{Desired architecture for an interpretable neural network representing $\mathbb{E}[y]=f_1(x_1)+f_2(x_2)+f_3(x_3)+f_{1,2}(x_1,x_2)$. The circles represent single nodes and the squares represent combinations of nodes.}
  \label{fig:idea_network}
\end{figure}

Neural networks also have the capacity to learn complex dependencies from data and can capture non-linear relationships between input and output variables, as well as interactions between variables. Contrary to the additive model, but similar to PPR, the results do not easily transfer to an interpretable model, since there will be links between all inputs and outputs through the inner layers. 
For a one-hidden-layer neural network, the neural network model resembles PPR, with the weight matrix elements (links) corresponding to the projection parameters $\bm{\beta_j}$ and an activation function $\sigma$ that generates $f_j \propto \sigma(\bm{\beta_j}^\top \bm{x})$. 
Now, we recognize that some neural network parameters (links) contribute more to the prediction than others, and there is reason to believe that if the less important links are removed from the network architecture, an interpretable structure would appear, as sketched in Figure \ref{fig:idea_network}. 
That is, sparse weights (projections) will enhance interpretability. Furthermore, by utilizing \emph{multiple hidden nodes}, corresponding to squares in Figure \ref{fig:idea_network}, to obtain the additive components, a diverse set of data-adaptive $f_j$, with varying complexity, can be obtained. 

A popular approach to eliminate parameters from a model is $l_1$-penalization, hereafter referred to as lasso \citep{tibshirani1996regression}, which, together with its extensions such as the group lasso \citep{yuan2006model} and the adaptive lasso \citep{zou2006adaptive}, has been used in neural networks to reduce complexity and increase interpretability.
Group lasso is used to penalize all the weights in a predefined group, typically all the inputs or outputs from a node or, in the case of convolutional neural networks, a channel \citep{scardapane2017group,wang2018novel,lym2019prunetrain,zhang2019feature,tank2017interpretable,ainsworth2018oi}. An alternative to group lasso is to add extra parameters that are multiplied to all the weights in a group and then apply standard lasso to these parameters \citep{leclerc2018smallify,sun2017design,wang2018interpret,vaughan2018explainable}. 
Adaptive lasso, where each parameter is assigned a unique penalty, has been used 
by \citet{xu2016adaptive} and \citet{qiao2018adaptive} to improve estimation stability, but without focus on interpretability.
Works cited above that focus on interpretability are those by \citet{tank2017interpretable}, where Granger causality between time series is restrained; \citet{ainsworth2018oi}, where multiple decoders are connected only to a subset of the latent variables generated by variational autoencoders; and, most relevant for our work, by \citet{wang2018interpret}, Distillation Guided Routing (DGR) and \citet{vaughan2018explainable}, Explainable Neural Networks (xNN).

In DGR, unimportant channels in a trained convolutional neural network are removed by multiplying the outputs from each channel with a lasso penalized parameter, balancing removal of channels with maintained performance. Since DGR removes entire channels, it can never remove unimportant links from important nodes. In Figure \ref{fig:idea_network} this corresponds to penalizing only the weights between the $f$'s and $y$.

In xNN, parallel neural networks with lasso penalized inputs are used to model each function in an additive model. xNN removes individual links, but only between the inputs and the, still black-box, parallel networks. In Figure \ref{fig:idea_network}, each $f$-box would correspond to a neural network of pre-defined capacity, where the inputs of the functions are chosen by the model, but where function complexities are pre-defined.

Here, we propose PrAda-net, a novel method that combines adaptive lasso for individual penalties of network parameters with proximal gradient descent. PrAda-net addresses two problems with lasso regularization of neural networks. First, lasso penalizes all model parameters equally, yet it is reasonable to assume that in an overparameterized neural network some weights contribute more to the final result than others, and should ideally be penalized less. Second, gradient descent algorithms cannot handle the non-differentiability of the lasso penalty at zero, something that has to be circumvented, e.g.\ by defining $\left.\frac{\partial |x|}{\partial x}\right|_{x=0}:=0$ or by using some smooth approximation, such as $|x| \approx \sqrt{x^2+\varepsilon}$. However, such approximations result in no parameters set exactly to zero, introducing the need for thresholding, which requires tuning as well. 

PrAda-net can be thought of as a generalization of xNN, where the number of hidden nodes to realize a function is adaptively chosen, depending on the function complexity. Therefore, it both improves the interpretability and reduces the number of parameters.
Compared to DGR, PrAda-net penalizes not only the outputs, but also the inputs, of the functions, resulting in a clear-cut identification of functional components. Compared to non-parametric, additive models, PrAda-net is completely data adaptive and neither the input, nor the form of the functional components, $f_j$, need to be pre-specified.

The rest of this paper is structured as follows: In Section \ref{sec:method}, we present the PrAda-net algorithm. In Section, \ref{sec:experiments} we illustrate the method on simulated data, comparing the predictive and interpretable performance of PrAda-net to that of standard lasso and DGR. We also apply PrAda-net to U.K.\ black smoke data. We compare the model components automatically selected by PrAda-net to the large-scale generalized additive model presented by \citet{wood2017generalized}. We show how the method is able to automatically select the network complexity to generate a concise, interpretable architecture, thus making it suitable for non-parametric additive modelling.
Finally, in Section \ref{sec:discussion}, we discuss our findings and possible future directions for research.

\section{Method}
\label{sec:method}
In this paper the following notation will be used: The input nodes of a neural network will be denoted by the vector $\bm{x}$, with elements $x_j$, and the single output node will be denoted by $y$.  Functions realized by hidden nodes of the network will be denoted by $f_j(\bm{x}|\bm{\theta})$ where $\bm{\theta}$ denotes the vector of network parameters, which will interchangeably be referred to as parameters, weights and links. Usually, $\bm{\theta}$ will be implicit, writing only $f_j(\bm{x})$.
A hat over a parameter or a function, e.g.\ $\hat{y}$, means that this is the reconstructed parameter or function inferred by the network; when it is clear from the context, the hat will be omitted.

As sketched in Figure \ref{fig:idea_network}, the architecture of a sparse neural network can be interpreted as an additive model, where each function is defined as the set of hidden nodes connected to the same set of input nodes.
In order to use lasso to obtain such an architecture, the two limitations stated above have to be overcome. First, lasso penalizes all parameters equally, while ideally, if the true model were known, one would want a very small penalty on the true model parameters, and a large penalty on parameters that should not be included. In the neural network setting, it is better to think of parameters as "relevant" or "important" rather than "true", but the rationale for data-adaptively eliminating parameters still stands. It was shown by \citet{zou2006adaptive} that under certain conditions, standard lasso is not consistent and as a remedy the adaptive lasso was proposed. In adaptive lasso, each parameter obtains an individual penalty based on its ordinary least square (OLS) estimate, penalizing parameters with small OLS values more. 

Second, lasso regularized models trained with gradient descent do not obtain explicit zeroes, which is a consequence of the non-uniqueness of the derivative of the lasso penalty at zero.  Proximal gradient descent \citep{rockafellar1976monotone}, on the other hand, leverages on this non-uniqueness and obtains exact zeros. Like standard gradient descent, proximal gradient descent is an iterative method, with the difference that the non-differentiable parts of the objective function are handled in an additional, proximal, step. If the objective function, $l(\bm{\theta})$, can be decomposed into $l(\bm{\theta}) = g(\bm{\theta}) + h(\bm{\theta})$, where $g$ is differentiable (typically a reconstruction error) and $h$ is not (typically a lasso regularization term), then a standard gradient descent step followed by a proximal gradient descent step is defined as
\begin{equation*}
\bm{\theta}^{t+1} = \textrm{prox}_{\alpha h}(\bm{\theta}^t-\alpha\nabla g(\bm{\theta}^t))
\label{eq:prox_grad_desc}
\end{equation*}
where $\alpha$ is a step size and prox is the proximal operator, that depends on $h$. In the case of lasso with penalty parameter $\lambda$, when $h(\bm{\theta}) = \lambda||\bm{\theta}||_1 = \lambda\sum_j |\theta_j|$, the proximal operator decomposes component-wise and is, with $\eta_j:= (\theta_j-\alpha\nabla g(\theta_j))$ denoting the output of the standard gradient descent step, 
\begin{equation*}
\textrm{prox}_{\alpha h}(\eta_j) = \textrm{sign}(\eta_j)\cdot \textrm{max}(|\eta_j| - \alpha \lambda, 0).
\end{equation*}
I.e., each $\eta_j$ is adaptively shrunk towards zero, and once it changes sign it is set to exactly zero.

\subsection{The PrAda-net Algorithm}
PrAda-net builds on the assumption that once a neural network is trained, even if all nodes in two subsequent layers are connected, some of the weights are more important than others, and that these important weights are larger in absolute value than the less important ones. Thus, adding an adaptive lasso penalty, i.e.\ an individual penalty based on the current value of the weight, and continue training will penalize the small, unimportant, weights more than the important ones, ideally setting all unimportant weights to zero, while penalizing the important weights less, thus reducing bias. Adding the adaptive lasso penalty changes the objective function as below:
\begin{equation}
\underset{\bm{\theta}}{\textrm{argmin}}\ l(\bm{x}|\bm{\theta}) \implies \underset{\bm{\theta}}{\textrm{argmin}}\left( l(\bm{x}|\bm{\theta}) + \lambda \sum_j \frac{|\theta_j|}{|\hat{\theta}_j|^\gamma}\right),
\label{eq:adaptive_lasso}
\end{equation}
where $\hat{\theta}_j$ denotes the value of the weight just before adding the penalty and $\gamma>0$ is a tunable parameter that we here set to 2, which is common practice in adaptive lasso \citep{zou2006adaptive}.

In order to get exact zeros one would like to use proximal gradient descent for this phase of the training. However, proximal gradient descent is not compatible with momentum based optimization algorithms, such as Adam \citep{kingma2014adam}. Furthermore, the choice of the step size $\alpha$, is not as obvious as in Adam, where the default parameters have been shown to work very well in most cases. We circumvent these issues by training PrAda-net in three stages: In the first stage, the network is trained with a standard optimizer, but without the lasso penalty, until convergence or some other stopping criterion. In the second stage, the adaptive lasso penalty is turned on, using the same optimizer as in the first stage, i.e. not utilizing proximal gradient descent, thus obtaining some weights close to, but distinct from, zero. Before the third stage the adaptive lasso penalty is updated based on the current weight values and then proximal gradient descent is used. Since some weights are now very close to zero, even with small values of $\alpha$ and $\lambda$, these weights will be heavily penalized and proximal gradient descent converges in relatively few iterations. For stage three, we thus used $10^{-5}$ for both $\alpha$ and $\lambda$, since the algorithm provided a stable behaviour around this value.
PrAda-net is summarized in Algorithm \ref{alg:adaptive_lasso}. The more crucial selection of tuning parameters in stage 2 is discussed separately in section \ref{sec:reg_strength}.

\begin{algorithm}
\caption{PrAda-net}
\begin{algorithmic}[1]
 \State Train the neural network until convergence or other stopping criterion, with optimizer of choice.
 \State Add an adaptive lasso penalty to each weight according to Equation \ref{eq:adaptive_lasso}.
Continue training with the chosen optimizer.
 \State Update the adaptive lasso penalty. Continue training with proximal gradient descent.
 
\end{algorithmic}
\label{alg:adaptive_lasso}
\end{algorithm}

\subsection{Choosing the Lasso Penalty Parameter}
\label{sec:reg_strength}
Choosing the optimal regularization parameter, $\lambda^*$, is a trade off between interpretability and model performance. When performance is the sole priority, the optimal regularization is often chosen so that it minimizes the mean test error across multiple splits of the data into training and test sets (cross-validation).
Here, we are willing to sacrifice some performance to gain interpretability and use a similar approach as in standard lasso regression packages \citep{friedman2010regularization}. We thus choose our $\lambda^*$ as the largest one of those whose resulting mean test error is within two standard deviations of the minimum test error across all $\lambda$. Figure \ref{fig:select_lbda} in Section \ref{sec:legendre} illustrates this.

\subsection{Choosing Stopping Criterion}
The most obvious stopping criterion for all three stages in PrAda-net is to train until the performance on the test set stops improving. However, it is known that early stopping can work as a regularizer, which is sometimes desirable. Nonetheless, our experience is that using PrAda-net with early stopping is inferior compared to training until convergence. Too early stopping in the first stage tends to result in $\hat{\theta}_j$ in Equation \ref{eq:adaptive_lasso} to exhibit an insufficient spread of magnitude for the adaptive lasso penalty have an impact. This thus reduces the ability of PrAda-net to generate compact model components. Too early stopping in one of the two last stages tends to result in unnecessarily complex models, using more hidden nodes than necessary. In all our experiment full convergence was used as the stopping criterion.

\subsection{Identifying Linear Functions}
As activation function for the hidden nodes, tanh was used. The slope of tanh varies smoothly between 0 and 1 on its support, which makes it suitable to approximate an unknown, smooth function. Another popular activation function is the rectified linear unit, $\max(0,x)$. In contrast to tanh, this function is piecewise linear, and using it for approximating a smooth function would in general require more hidden nodes, or a rougher approximation, than when using tanh. 

However, as can be seen from its Maclaurin expansion $\tanh(x)=x+\mathcal{O}(x^3)$, is almost linear if $x$ is small in absolute value. For $x_1$ close to zero, $\textrm{tanh}(x_1+x_2) \approx x_1+\textrm{tanh}(x_2)$, and therefore, if the output is linear in some input variable ($x_1$ in this example), no separate hidden node is needed to model this linear function. Instead, the function can be incorporated into some other node, thus increasing compactness but decreasing interpretability. Ideally, a separate penalty or regularization technique should be included to avoid this, but since it is far from obvious how this penalty should be formulated while retaining computational feasibility, instead the following postprocessing step was added: For each identified function, the partial derivatives with respect to all its inputs were calculated and if, for some input, the derivative was close to constant, i.e.\ with variance lower than some  $\sigma^2_{\textrm{max}}$, that input was removed from the function and put into a new, linear, function, realized by a new node. $\sigma^2_{\textrm{max}}$ was calibrated on synthetic data to 0.01. Since for standardized data, $\sigma^2_{\textrm{max}}$ translates between data sets, this value was consistently used throughout the paper.

\section{Experiments}
\label{sec:experiments}
All experiments in this section were done on neural networks with one hidden layer, with tanh as activation function, and one output node with a linear activation function. The mean squared error was used as the loss function and the Adam optimizer was used in the two first stages of the PrAda-net algorithm. The data was randomly split into 90 \% training and 10 \% testing data. When choosing $\lambda^*$, 20 random splits of the data were made for each $\lambda$ value. In order to escape suboptimal minima, five different initializations were used for each training. All computations were done on a NVIDIA V100, 32GB, GPU.

Three experiments were performed. In the first experiment, synthetic data was generated from the first four Legendre polynomials; in the second experiment, we used a data set with black smoke levels in the U.K.\ recently analyzed by using a large-scale generalized additive model by \cite{wood2017generalized}; and in the third experiment synthetic data was generated from the model inferred by PrAda-net on the U.K.\ black smoke data.

\subsection{Legendre Polynomials}
\label{sec:legendre}
Our first experiment was done on the sum of the first four Legendre polynomials, which are orthogonal on the interval $[-1,1]$ and are given by
\begin{equation}
\begin{aligned}
&P_1(x)=x\\
&P_2(x)=\frac{1}{2}(3x^2-1)\\
&P_3(x)=\frac{1}{2}(5x^3-3x)\\
&P_4(x) = \frac{1}{8}(34x^4 - 30x^2 + 3).
\end{aligned}
\label{eq:leg}
\end{equation} 
For the simulations, five random variables, $x_1,\dots x_5$, were generated from $\mathcal{U}[-1,1]$, each with 1000 realizations, resulting in a $1000 \times 5$ $x$-matrix, from which a $1000 \times 1$ $y$-matrix was created according to
\begin{equation}
\begin{aligned}
&y=P_1(x_1) + P_2(x_2) + P_3(x_3) + P_4(x_4) + \varepsilon\\
\end{aligned}
\label{eq:y_leg}
\end{equation} 
where $\varepsilon\sim \mathcal{N}(0,0.1^2)$ is added noise. Note that $y$ is independent of $x_5$. A neural network with five input nodes and 50 hidden nodes was trained by applying Algorithm \ref{alg:adaptive_lasso} with $\lambda^*$ chosen as explained in Section \ref{sec:reg_strength}.

\begin{figure*}
     \centering
     \begin{subfigure}[b]{0.39\textwidth}
         \centering
         \includegraphics[width=\textwidth]{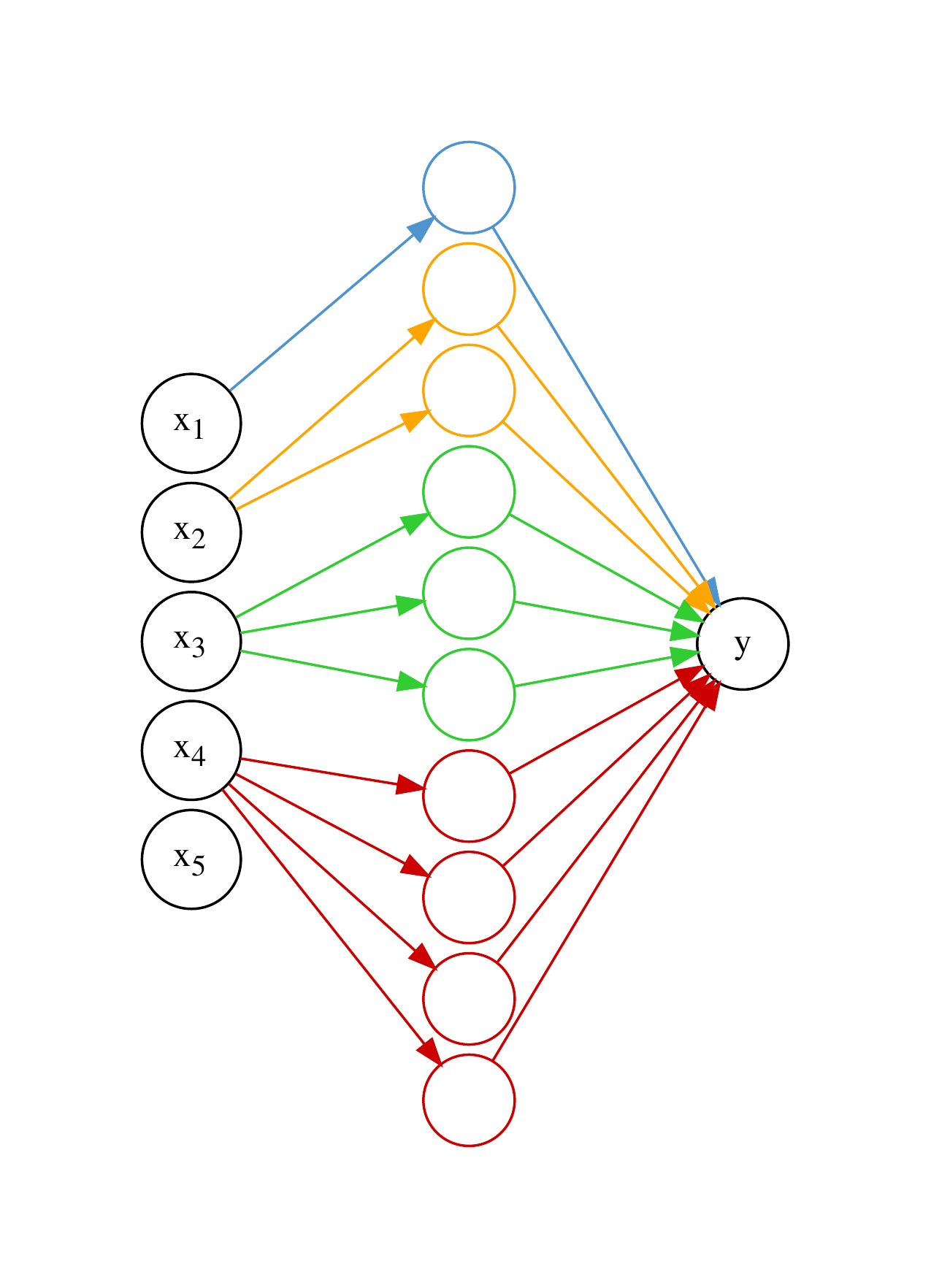}
         \caption{}
         \label{fig:leg_graph}
     \end{subfigure}
     \hfill
     \begin{subfigure}[b]{0.6\textwidth}
         \centering
         \includegraphics[width=\textwidth]{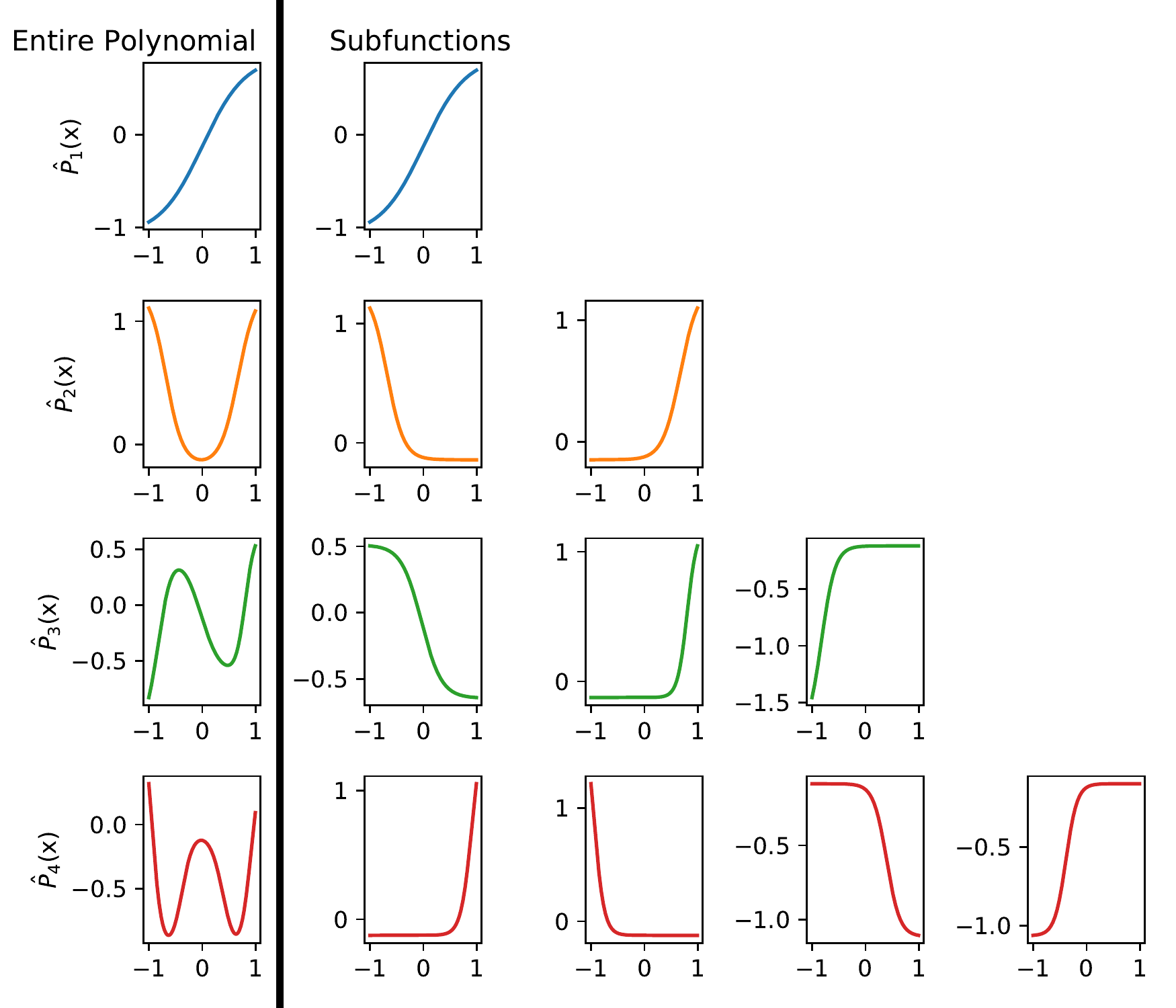}
         \caption{}
         \label{fig:plot_legendre}
     \end{subfigure}
     \hfill
     \begin{subfigure}[b]{0.6\textwidth}
         \centering
         \includegraphics[width=\textwidth]{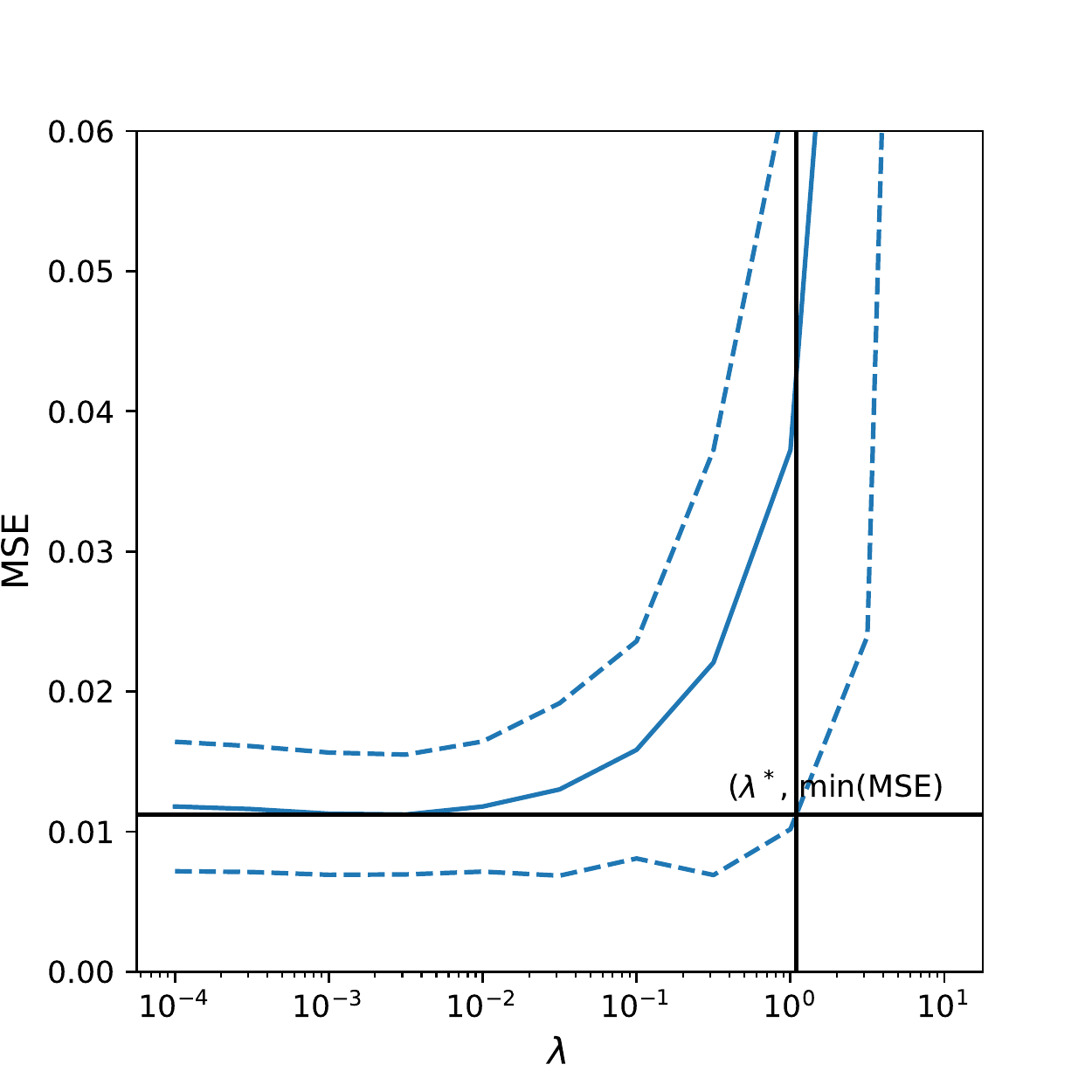}
         \caption{}
         \label{fig:select_lbda}
     \end{subfigure}
     \caption{Identifying the sum of the four first Legendre polynomials. \ref{fig:leg_graph} shows the inferred reduced neural network, with the hidden nodes and links color coded according to the identified functions.
In \ref{fig:plot_legendre}, the leftmost column shows, from top to bottom, $\hat{P}_1(x_1)$, $\hat{P}_2(x_2)$, $\hat{P}_3(x_3)$ and $\hat{P}_4(x_4)$, while the other columns show their decompositions into the subfunctions realized by single hidden nodes. \ref{fig:select_lbda} shows, the mean $\pm$ two standard deviations of the test error obtained by 20 bootstrap runs for each $\lambda$. $\lambda^*$ is chosen as the value where the lower interval intersects the lowest mean value.}
     \label{fig:leg_res}
\end{figure*}

The results are shown in Figure \ref{fig:leg_res}. As seen in \ref{fig:leg_graph}, out of the original 50 hidden nodes, only 10 are used in the final architecture and the hidden units are split into four different functions, each with only one $x_j$ as input. Figure \ref{fig:plot_legendre} shows the reconstructed $\hat{P}_j(x)$'s (left-most panels) and also how the model decomposes each $\hat{P}_j(x)$ into a sum of up to four subfunctions, each represented by one node in the hidden layer. Figure \ref{fig:select_lbda} demonstrates the penalty parameter selection for this simulation.

We compared the test error and the variable importance, measured using saliency maps \citep{simonyan2013deep} for PrAda-net, standard lasso and DGR. To make standard lasso more competitive, we did not apply lasso regularization from the start. Instead, we utilized the PrAda-net algorithm with the same penalty for all links, i.e.\ $\gamma=0$ in Equation \ref{eq:adaptive_lasso}. DGR was adapted to feedforward regression networks by looking at nodes instead of channels and by using the squared error instead of the cross-entropy loss. 

For saliency maps, the importance of variable $j$, evaluated at $\bm{x}$, is defined as $I_j(\bm{x}) :=\left|\frac{\partial f(\bm{x})}{\partial x_j}\right|$, which in our case reduces to $I_j(\bm{x}) =\left|\frac{\partial P_j(x_j)}{\partial x_j}\right|$. Since for non-linear functions the value of the derivative depends on $\bm{x}$, in order to get a global importance measure, we averaged over all $\bm{x}$'s in the test set $\mathcal{X}$, $I_j := \frac 1 {|\mathcal{X}|}\sum_\mathcal{X} I_j(\bm{x})$. This can be seen as a Monte Carlo integral, meaning that as $|\mathcal{X}| \to \infty$, $I_j \to \frac{1}{2}\int_{-1}^{1}|\frac{\partial P_j(x_j)}{\partial x_j}|dx_j$, where $\frac{1}{2}$ is the probability density function of the uniform distribution that $x_j$ is sampled from. Thus, for large sample sizes, the true value of the variable importance is given by the analytical integral.

50 runs with different noise realizations were performed with PrAda-net penalized with $\lambda^*$, and the other two models penalized to equal complexity. In the case of standard lasso, the regularization was chosen to obtain the same number of parameters in both models, while for DGR this would be an unfair comparison, since penalization is done at node level. Instead, for DGR, regularization was chosen to have the same number of nodes as PrAda-net, allowing it to use many more parameters than PrAda-net and standard lasso.

The results are summarized in Table \ref{tab:var_imp}. PrAda-net outperforms both standard lasso and DGR, both in terms of test error and variable importance. This effect is especially notable for the higher order functions. While all methods did fairly well at identifying the noise variable as such, PrAda-net was the only method to do so in all 50 runs.

\begin{table*}
\centering
\caption{Legendre polynomials: Mean and one standard deviation of test error and estimated variable importance ($\frac 1 {|\mathcal{X}|}\sum_\mathcal{X}\left|\frac{\partial f(x)}{\partial x_j}\right|$) together with the true asymptotic variable importance ($\frac{1}{2}\int_{-1}^{1}|\frac{\partial P_j(x_j)}{\partial x_j}|dx_j$).}
\label{tab:var_imp}
\begin{tabular}{ |c|c|c|c|c|c|c| } 
\hline
& True Var.Imp. &PrAda-net&Standard Lasso&DGR\\
\hline
Test error & - & 0.04 $\pm$ 0.02 & 0.13 $\pm$ 0.03 & 0.16 $\pm$ 0.03\\
\hline
$x_1$ & 1 & 0.75 $\pm$ 0.33 & 0.69 $\pm$ 0.43 & 0.63 $\pm$ 0.09\\
$x_2$ & 1.5 & 1.26 $\pm$ 0.09 & 1.21 $\pm$ 0.05 & 0.65 $\pm$ 0.12\\
$x_3$ & 1.89 & 1.42 $\pm$ 0.16 & 0.96 $\pm$ 0.11 & 0.32 $\pm$ 0.12\\
$x_4$ & 2.23 & 1.72 $\pm$ 0.3 & 0.43 $\pm$ 0.24 & 0.25 $\pm$ 0.08\\
$x_5$ & 0 & 0.0 $\pm$ 0.0 & 0.01 $\pm$ 0.01 & 0.07 $\pm$ 0.04\\

\hline
\end{tabular}
\end{table*}

We also investigated how well the algorithms could identify the four different functions in the true model. In this comparison, DGR was left out since all input nodes are connected to all nodes in the hidden layer, resulting in only $f(x_1, x_2, x_3, x_4, x_5)$ being identified. Standard lasso was penalized with two different strengths, one to get the same number of parameters as for PrAda-net and one to get the same number of functions.

Table \ref{tab:sub_fcts} shows that PrAda-net is able to identify the true functions in the model as well as their complexity, measured in number of nodes. That is, PrAda-net assigns more nodes to the higher order polynomials. The true functions are identified much more often by PrAda-net than by standard lasso. Standard lasso with the same complexity erroneously includes higher-order interaction terms (e.g.\ $f(x_1,x_2,x_3)$) rather than increasing the presence or function complexity of the main effect terms (e.g.\ $f(x_2)$). Standard lasso with the same number of functions successfully identifies $f_1(x_1)$ and $f_2(x_2)$ but fails to identify $f_3(x_3)$ and $f_4(x_4)$.

\begin{table*}
\centering
\caption{Legendre polynomials: Presence (proportion of simulations where function is identified) and average complexity of functions across 50 simulation runs. The true model components (main effects) are marked with an asterisk. Complexity of identified functions is measured in number of nodes.}
\label{tab:sub_fcts}
\begin{tabular}{ |l|c|c|c|c|c|c| } 
\hline
         & \multicolumn{2}{c|}{PrAda-net} & \multicolumn{2}{c|}{Standard Lasso}  & \multicolumn{2}{c|}{Standard Lasso}\\
         & \multicolumn{2}{c|}{ } & \multicolumn{2}{c|}{(same number}  & \multicolumn{2}{c|}{(same}\\
         & \multicolumn{2}{c|}{ } & \multicolumn{2}{c|}{(of functions)}  & \multicolumn{2}{c|}{complexity) }\\
\hline
Function              &Pres-& Comp-  & Pres- & Comp- & Pres-  & Comp-\\
                      &ence & lexity & ence  & lexity& ence   & lexity\\
\hline
$f(x_1)^*$ & 1.0 & 1.0 & 1.0 & 1.72 & 0.12 & 1.0\\
$f(x_1,x_2)$ & - & - & - & - & 0.12 & 1.17\\
$f(x_1,x_2,x_3)$ & - & - & - & - & 0.14 & 1.0\\
$f(x_1,x_2,x_3,x_4)$ & - & - & - & - & 0.82 & 1.46\\
$f(x_1,x_2,x_3,x_4,x_5)$ & - & - & - & - & 0.44 & 1.0\\
$f(x_1,x_2,x_3,x_5)$ & - & - & - & - & 0.02 & 1.0\\
$f(x_1,x_3)$ & - & - & - & - & 0.96 & 1.06\\
$f(x_1,x_3,x_4)$ & - & - & - & - & 0.64 & 1.25\\
$f(x_1,x_3,x_4,x_5)$ & - & - & - & - & 0.06 & 1.0\\
$f(x_1,x_4)$ & - & - & - & - & 0.42 & 1.0\\
$f(x_2)^*$ & 1.0 & 1.96 & 0.92 & 2.0 & 0.14 & 1.14\\
$f(x_2,x_3)$ & - & - & 0.08 & 2.0 & 0.02 & 1.0\\
$f(x_2,x_3,x_4)$ & - & - & - & - & 0.04 & 1.0\\
$f(x_2,x_3,x_5)$ & - & - & - & - & 0.02 & 1.0\\
$f(x_2,x_4)$ & 0.04 & 1.0 & - & - & 0.08 & 1.0\\
$f(x_2,x_5)$ & - & - & - & - & 0.02 & 1.0\\
$f(x_3)^*$ & 1.0 & 2.62 & 0.1 & 1.6 & 0.2 & 1.6\\
$f(x_3,x_4)$ & 0.04 & 1.0 & 0.02 & 1.0 & 0.1 & 1.2\\
$f(x_4)^*$ & 1.0 & 3.98 & 0.06 & 1.0 & 0.96 & 1.83\\
\hline
\end{tabular}
\end{table*}

\subsection{Black Smoke Data}
To illustrate the usefulness of PrAda-net on real, large-scale data with a complex structure, we applied it to the U.K.\ black smoke network daily data set, analyzed by \citep{wood2017generalized} and available on the main author's homepage\footnote{\url{https://www.maths.ed.ac.uk/~swood34/data/black_smoke.RData}}. The data set, collected over four decades, is massive, comprising 10 million observations of air pollution data (daily concentration of black smoke particulates in $\mu g m^{-3}$) measured at more than 2000 monitoring stations.
In addition to the pollution data, \texttt{bs}, the following eleven covariates are available: year, \texttt{y}; day of year, \texttt{doy}; day of week, \texttt{dow}; location as kilometers east, \texttt{e} and north, \texttt{n}; height, \texttt{h}; cubic root transformed rainfall, \texttt{r}; daily minimum and maximum temperature \texttt{T}$^\texttt{0}$ and \texttt{T}$^\texttt{1}$; and mean temperatures the previous two days, $\bar{\texttt{T1}}$ and $\bar{\texttt{T2}}$.

\citet{wood2017generalized} first perform a separate spatial modeling and then a separate temporal modeling of the data to propose some candidate model components. In a final modeling step, the components are combined through a generalized additive modeling approach, including interactions between the proposed model components. Specifically, the log transformed black smoke level is modelled as a sum of fourteen functions, each containing up to three of the eleven covariates. In addition to these functions, the model includes an offset for each station site type $k$ (rural, industrial, commercial, city/town center or mixed), $\alpha_k$; a station specific random effect, $b$; and a time correlation model for the error term following an AR process, $e$. The final model is given by 
\begin{equation}
\label{eq:wood}
\begin{aligned}
 \log(\texttt{bs})=&f_{1}(\texttt{y})+f_{2}(\texttt{doy})+f_{3}(\texttt{dow})+f_{4}(\texttt{y},\texttt{doy})+f_{5}(\texttt{y},\texttt{dow})\\
&+f_{6}(\texttt{doy},\texttt{dow}) +f_{7}(\texttt{n},\texttt{e})+f_{8}(\texttt{n},\texttt{e},\texttt{y})+f_{9}(\texttt{n},\texttt{e},\texttt{doy})\\
&+f_{10}(\texttt{n},\texttt{e},\texttt{dow})+f_{11}(\texttt{h})+f_{12}(\texttt{T}^\texttt{0},\texttt{T}^\texttt{1})+f_{13}(\bar{\texttt{T1}},\bar{\texttt{T2}})\\
&+f_{14}(\texttt{r})+\alpha_k+b+e.
\end{aligned}
\end{equation}

For functional forms, cubic splines were used for the temporal components, thin-plate splines for the spatial components and tensor product smoothers for the interactions. In summary, this analysis builds on a substantial and non-trivial preliminary screening and modeling of the data.

In contrast, our approach is to let PrAda-net with 100 hidden nodes automatically decide the format of the additive model functions. To make the results reasonably comparable we use a penalization parameter for PrAda-net to obtain, approximately, the same number of functions as in Equation \ref{eq:wood}. Note, however, that this generally produces slightly fewer functions for PrAda-net than the model by \citet{wood2017generalized} since we do not post-process to combine functions that are decomposed into main and interaction terms by PrAda-net. E.g.\ the single function $f(e,n)$ in \citet{wood2017generalized} is decomposed into two main effects, $f(e)$ and $f(n)$, and one interaction effect, $f(e,n)$, by PrAda-net (see Equation \ref{eq:prada_medoid} and Table \ref{tab:bs_deduced}) and is thus realized by three functions. It is non-trivial to conduct such post-processing for all possible main and higher order interactions so we elect to choose a conservative model for PrAda-net. 

We ran PrAda-net for 20 different random splits of the data which resulted in, in total, 52 different functions, 6 of which contained more than three covariates, see Table \ref{tab:bs_deduced}. The medoid model, where the distance between models was measured by the Jaccard distance between the set of included functions, is given by
\begin{equation}
\begin{aligned}
\log(\texttt{bs})_{\textrm{PrAda}}=&\underbrace{f_{1}(\texttt{y})}_{10}+ \underbrace{f_{2}(\texttt{doy})}_{4}+ \underbrace{f_{3}(\texttt{dow})}_{2}+ \underbrace{f_{4}(\texttt{T}^\texttt{0})}_{1}+ \underbrace{f_{5}(\texttt{T}^\texttt{1})}_{2}
+ \underbrace{f_{6}(\bar{\texttt{T2}})}_{1}\\
&+ \underbrace{f_{7}(\texttt{r})}_{1} + \underbrace{f_{8}(\texttt{e})}_{3} + \underbrace{f_{9}(\texttt{n})}_{8}+\underbrace{f_{10}(\texttt{h})}_{2}+ \underbrace{f_{11}(\texttt{doy},\texttt{T}^\texttt{1})}_{2}+ \underbrace{f_{12}(\texttt{e},\texttt{n})}_{12}\\
&+ \underbrace{f_{13}(\texttt{e},\texttt{h})}_{1} + \underbrace{f_{14}(\texttt{n},\texttt{h})}_{1}  + \underbrace{f_{15}(\texttt{T}^\texttt{0},\texttt{T}^\texttt{1},\bar{\texttt{T1}})}_{1}+ \underbrace{f_{16}(\texttt{e},\texttt{n},\texttt{h})}_{3}\\
&+\alpha_k+b+e,
\label{eq:prada_medoid}
\end{aligned}
\end{equation}
where the number under a function denotes its complexity, measured in the number of hidden nodes used to realize it. Six of the sixteen functions in the PrAda-net medoid model overlap with the manually selected functions in \citet{wood2017generalized}. 
However, the pre-identified interaction terms between temporal covariates and between 
temporal and spatial covariates are not selected by PrAda-net, which, on the other hand, selects a more complex spatial interaction, $f_{16}(\texttt{e},\texttt{n},\texttt{h})$.

For standard lasso, the 20 data splits resulted in 143 different functions, 103 of which contained more than 3 covariates, see Table \ref{tab:bs_deduced}, with medoid function
\begin{equation}
\begin{aligned}
\log(\texttt{bs})_{\textrm{Lasso}}&=\underbrace{f_{1}(\texttt{dow})}_{2}+ \underbrace{f_{2}(\texttt{e})}_{1}+ \underbrace{f_{3}(\texttt{n})}_{1}+ \underbrace{f_{4}(\texttt{h})}_{1}+ \underbrace{f_{5}(\texttt{doy},\bar{\texttt{T2}})}_{1}+ \underbrace{f_{6}(\texttt{e},\texttt{n})}_{1}\\
&+ \underbrace{f_{7}(\texttt{r},\texttt{h})}_{1} + \underbrace{f_{8}(\texttt{n},\texttt{h})}_{1}+ \underbrace{f_{9}(\texttt{y},\texttt{r},\texttt{e})}_{1} + \underbrace{f_{10}(\texttt{y},\texttt{e},\texttt{n})}_{1}\\
&+ \underbrace{f_{11}(\texttt{y},\texttt{doy},\texttt{T}^\texttt{1},\texttt{r})}_{1}+\underbrace{f_{12}(\texttt{y},\texttt{T}^\texttt{1},\bar{\texttt{T1}},\texttt{r})}_{1}+ \underbrace{f_{13}(\texttt{y},\texttt{e},\texttt{n},\texttt{h})}_{1}\\
& + \underbrace{f_{14}(\texttt{T}^\texttt{0},\bar{\texttt{T1}},\bar{\texttt{T2}},\texttt{h})}_{1}+\underbrace{f_{15}(\texttt{T}^\texttt{1},\texttt{r},\texttt{n},\texttt{h})}_{1} +\underbrace{f_{16}(\texttt{T}^\texttt{0},\texttt{T}^\texttt{1},\texttt{e},\texttt{n},\texttt{h})}_{1}\\
& + \underbrace{f_{17}(\texttt{y},\texttt{doy},\texttt{T}^\texttt{0},\texttt{T}^\texttt{1},\texttt{r},\texttt{h})}_{1} + \underbrace{f_{18}(\texttt{y},\texttt{dow},\bar{\texttt{T1}},\texttt{r},\texttt{n},\texttt{h})}_{1}+\alpha_k+b+e.
\label{eq:lasso_medoid}
\end{aligned}
\end{equation}
Compared to standard lasso, PrAda-net has higher precision, i.e.\ it detects fewer unique functions more frequently, indicating a more stable selection performance. This is confirmed by the medoid model being more similar to the other models for PrAda-net than for standard lasso. The average Jaccard similarity of the PrAda-net medoid model is 0.63, compared with 0.18 for standard lasso. PrAda-net also identifies functions containing fewer covariates (order of interaction) thus resulting in more interpretable models compared to standard lasso.
All the $2\cdot 20$ models had an explained variance of $R^2\approx 0.79$, which is the same as reported by \citet{wood2017generalized}.

\begin{table*}
\centering
\caption{Identified functions in 20 bootstrap runs using PrAda-net and standard lasso. Only the functions present in more than half of the runs are shown.}
\label{tab:bs_deduced}
\begin{tabular}[t]{|l|l|l|l|l|} 
\hline
& \multicolumn{2}{l|}{PrAda-net} & \multicolumn{2}{l|}{Standard lasso}\\
\hline
\makecell{Identified functions\\in total} & \multicolumn{2}{l|}{52} & \multicolumn{2}{l|}{143}\\
\hline
\makecell{Functions with more\\than 3 covariates} & \multicolumn{2}{l|}{6} & \multicolumn{2}{l|}{103}\\
\hline
\multirow{3}{*}{\makecell{Most frequent functions\\with frequencies}} 

& $f(\texttt{doy})$ & 1.0 & $f(\texttt{e},\texttt{n})$ & 0.8\\
& $f(\texttt{y})$ & 1.0 & $f(\texttt{dow})$ & 0.8\\
& $f(\texttt{T}^\texttt{1})$ & 1.0 & $f(\texttt{n})$ & 0.7\\
& $f(\texttt{n})$ & 1.0 & $f(\texttt{n},\texttt{h})$ & 0.6\\
& $f(\texttt{e})$ & 1.0 & $f(\texttt{h})$ & 0.55\\
& $f(\texttt{e},\texttt{n},\texttt{h})$ & 1.0 & $f(\texttt{e})$ & 0.55\\
& $f(\texttt{e},\texttt{n})$ & 1.0 & &\\
& $f(\texttt{dow})$ & 0.95 & &\\
& $f(\texttt{h})$ & 0.9 & &\\
& $f(\texttt{n},\texttt{h})$ & 0.85 & &\\
& $f(\texttt{e},\texttt{h})$ & 0.8 & &\\
& $f(\texttt{T}^\texttt{0})$ & 0.75 & &\\
& $f(\texttt{doy},\texttt{T}^\texttt{1})$ & 0.6 & &\\
& $f(\bar{\texttt{T2}})$ & 0.55 & &\\
\hline
\end{tabular}
\end{table*}

\begin{figure*}
  \centering
  \includegraphics[width=1.\textwidth]{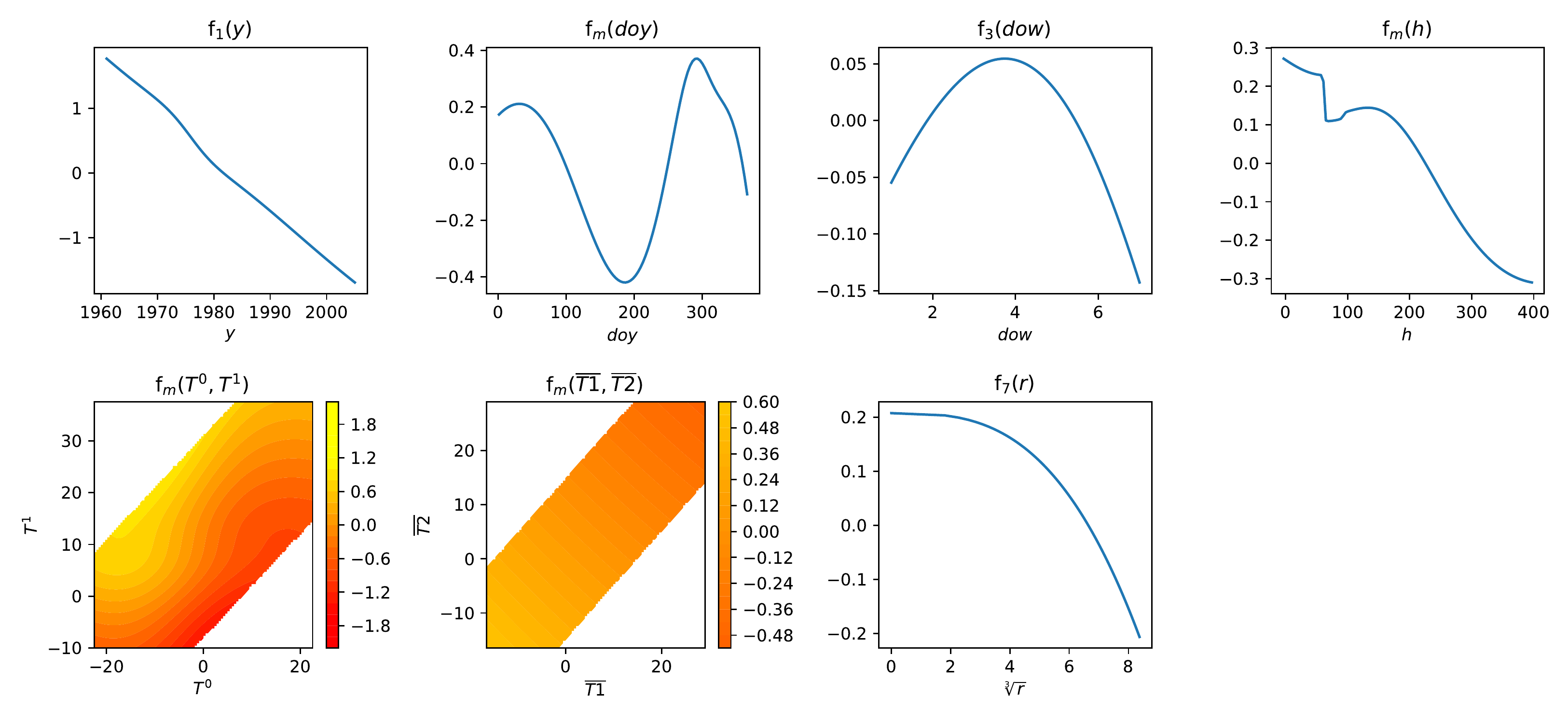}
  \caption{Functions identified by PrAda-net to be compared to functions by \citet{wood2017generalized}, Supplemental Figure 2, 
  { \url{https://www.tandfonline.com/doi/suppl/10.1080/01621459.2016.1195744}}. 
  Since day of year, height and the temperature covariates are present in multiple functions in PrAda-net, in these cases the marginal function is displayed. Subscript numbers refer to functions in Equation \ref{eq:prada_medoid}, while subscript m denotes marginal function.}
  \label{fig:bs_compare}
\end{figure*}

In Figure \ref{fig:bs_compare} we visually summarize a subset of the model components identified by PrAda-net in Equation \ref{eq:prada_medoid}. These same model components are presented in 
Figure 2 in the Supplemental material of \citet{wood2017generalized}\footnote{Available at \url{https://www.tandfonline.com/doi/suppl/10.1080/01621459.2016.1195744}}.  In the cases where PrAda-net splits the dependency of a covariate across multiple functions, the marginal function is plotted to allow for direct comparison with \cite{wood2017generalized}. The generated functions in Figure \ref{fig:bs_compare} bear a strong resemblance to those in Supplemental Figure 2 of \citet{wood2017generalized}, indicating that PrAda-net was able to automatically identify a highly similar model to the manually curated one. 
With the exception of height, $\texttt{h}$, PrAda-net tends to identify smoother functions than \citet{wood2017generalized}.

\subsection{Synthetic Black Smoke Data}
To test PrAda-net on a more complex model than the synthetic data in Section \ref{sec:legendre}, we generated data from the PrAda-net medoid model for the black smoke data. 10000 data points were sampled, according to $\mathcal{N}(0,\Sigma_S)$, where $\Sigma_S$ is the Spearman rank correlation matrix of the variables in the black smoke data set, resulting in a $10000 \times 11$ $x$-matrix, from which a $10000 \times 1$ $y$-matrix was created according to Equation \ref{eq:prada_medoid}, without $\alpha_k$, $b$, and $e$ and with added noise distributed as $\mathcal{N}(0,0.1^2)$. 50 simulations, with different noise realizations, were performed using a hidden layer with 100 units. For PrAda-net, $\lambda^*$ was chosen as described in Section \ref{sec:reg_strength} and for standard lasso, the regularization strength was chosen to obtain the same number of functions as for PrAda-net. The results are summarized in Table \ref{tab:sub_fcts_prada}, where all functions present in more than 20 \% of the simulations are shown and functions present in more than 50 \% of the simulations are marked with an asterisk.

PrAda-net identifies 14 of the 16 true model components together with 4 false positives, the corresponding numbers for standard lasso are 6 and 17, respectively. There is a tendency that more complex functions, i.e.\ functions consisting of more than one hidden node, are more easily identified. Except for $f_{5}(\texttt{T}^\texttt{1})$ and $f_{10}(\texttt{h})$, PrAda-net identifies all the functions realized by more than one hidden node in 100 \% of the simulations; for both algorithms all the falsely included functions also have a complexity close to one node. Apart from $f(\texttt{doy})$, PrAda-net seems to be closer to the true complexity than standard lasso.

\begin{table*}
\caption{Presence and average complexity of functions across 50 simulation runs for PrAda-net (left) and standard lasso (right), with true model components above the line and falsely included components below. Only functions present in more than 20 \% of the runs are presented, functions present in more than 50 \% of the runs are marked with an asterisk.}
\label{tab:sub_fcts_prada}
\small
\centering
\begin{tabular}{ |l|c|c| } 
\hline
\multicolumn{3}{|c|}{PrAda-net}\\
\hline
\makecell{Function}              &\makecell{Pres-\\ence}   &\makecell{Comp-\\lexity}\\
\hline
$f(\texttt{y})^*$ & 1.0 & 5.06\\
$f(\texttt{doy})^*$ & 1.0 & 20.92\\
$f(\texttt{dow})^*$ & 1.0 & 2.9\\
$f(\texttt{e})^*$ & 1.0 & 2.36\\
$f(\texttt{n})^*$ & 1.0 & 3.0\\
$f(\texttt{doy},\texttt{T}^\texttt{1})^*$ & 1.0 & 2.22\\
$f(\texttt{e},\texttt{n})^*$ & 1.0 & 13.44\\
$f(\texttt{n},\texttt{h})^*$ & 1.0 & 1.18\\
$f(\texttt{e},\texttt{n},\texttt{h})^*$ & 1.0 & 2.22\\
$f(\texttt{T}^\texttt{1})^*$ & 0.98 & 1.27\\
$f(\texttt{T}^\texttt{0})^*$ & 0.94 & 1.23\\
$f(\texttt{e},\texttt{h})^*$ & 0.72 & 1.44\\
$f(\texttt{h})$ & 0.48 & 1.04\\
$f(\texttt{r})$ & 0.22 & 1.0\\
\hline
$f(\texttt{doy},\bar{\texttt{T1}})$ & 0.4 & 1.05\\
$f(\texttt{doy},\texttt{e})$ & 0.3 & 1.07\\
$f(\texttt{dow},\texttt{e})$ & 0.24 & 1.17\\
$f(\texttt{T}^\texttt{1},\bar{\texttt{T1}})$ & 0.2 & 1.0\\
\hline
\multicolumn{3}{c}{ }\\
\multicolumn{3}{c}{ }\\
\multicolumn{3}{c}{ }\\
\multicolumn{3}{c}{ }\\
\multicolumn{3}{c}{ }\\
\end{tabular}
\begin{tabular}{ |l|c|c| } 
\hline
\multicolumn{3}{|c|}{Standard lasso}\\
\hline
Function              &\makecell{Pres-\\ence}   &\makecell{Comp-\\lexity}\\
\hline
$f(\texttt{y})^*$ & 1.0 & 15.8\\
$f(\texttt{e},\texttt{n})^*$ & 1.0 & 4.16\\
$f(\texttt{e},\texttt{n},\texttt{h})^*$ & 0.96 & 1.27\\
$f(\texttt{doy})^*$ & 0.82 & 1.76\\
$f(\texttt{n})^*$ & 0.82 & 1.37\\
$f(\texttt{h})^*$ & 0.56 & 1.0\\
\hline
$f(\texttt{dow},\texttt{e},\texttt{h})^*$ & 0.7 & 1.2\\
$f(\texttt{T}^\texttt{0},\texttt{e},\texttt{n},\texttt{h})^*$ & 0.58 & 1.03\\
$f(\texttt{doy},\texttt{T}^\texttt{1},\bar{\texttt{T2}},\texttt{e},\texttt{h})^*$ & 0.52 & 1.27\\
$f(\texttt{doy},\texttt{T}^\texttt{0},\texttt{T}^\texttt{1},\bar{\texttt{T2}},\texttt{r},\texttt{e},\texttt{n},\texttt{h})^*$ & 0.52 & 1.0\\
$f(\texttt{dow},\texttt{e})$ & 0.44 & 1.05\\
$f(\texttt{doy},\texttt{T}^\texttt{0},\texttt{T}^\texttt{1},\bar{\texttt{T1}},\bar{\texttt{T2}},\texttt{r},\texttt{e},\texttt{n},\texttt{h})$ & 0.34 & 1.18\\
$f(\texttt{doy},\texttt{T}^\texttt{0},\texttt{T}^\texttt{1},\texttt{e},\texttt{h})$ & 0.32 & 1.0\\
$f(\texttt{doy},\texttt{dow},\texttt{e},\texttt{h})$ & 0.3 & 1.07\\
$f(\texttt{doy},\texttt{T}^\texttt{1},\bar{\texttt{T2}},\texttt{e})$ & 0.3 & 1.0\\
$f(\texttt{doy},\bar{\texttt{T1}},\texttt{e},\texttt{h})$ & 0.3 & 1.0\\
$f(\texttt{doy},\texttt{T}^\texttt{0},\texttt{T}^\texttt{1},\bar{\texttt{T1}},\bar{\texttt{T2}},\texttt{h})$ & 0.3 & 1.0\\
$f(\texttt{doy},\texttt{T}^\texttt{0},\texttt{T}^\texttt{1},\bar{\texttt{T2}},\texttt{e},\texttt{h})$ & 0.3 & 1.2\\
$f(\texttt{T}^\texttt{0},\texttt{T}^\texttt{1},\bar{\texttt{T1}},\bar{\texttt{T2}},\texttt{r},\texttt{e},\texttt{n},\texttt{h})$ & 0.26 & 1.0\\
$f(\texttt{T}^\texttt{1},\texttt{n})$ & 0.24 & 1.0\\
$f(\texttt{T}^\texttt{0},\texttt{e},\texttt{n})$ & 0.24 & 1.0\\
$f(\texttt{doy},\texttt{T}^\texttt{0},\texttt{T}^\texttt{1},\bar{\texttt{T1}},\bar{\texttt{T2}},\texttt{r},\texttt{e},\texttt{h})$ & 0.24 & 1.0\\
$f(\texttt{doy},\texttt{T}^\texttt{1},\texttt{e},\texttt{h})$ & 0.22 & 1.0\\
\hline
\end{tabular}
\end{table*}

\begin{figure*}
  \centering
  \includegraphics[width=0.9\textwidth]{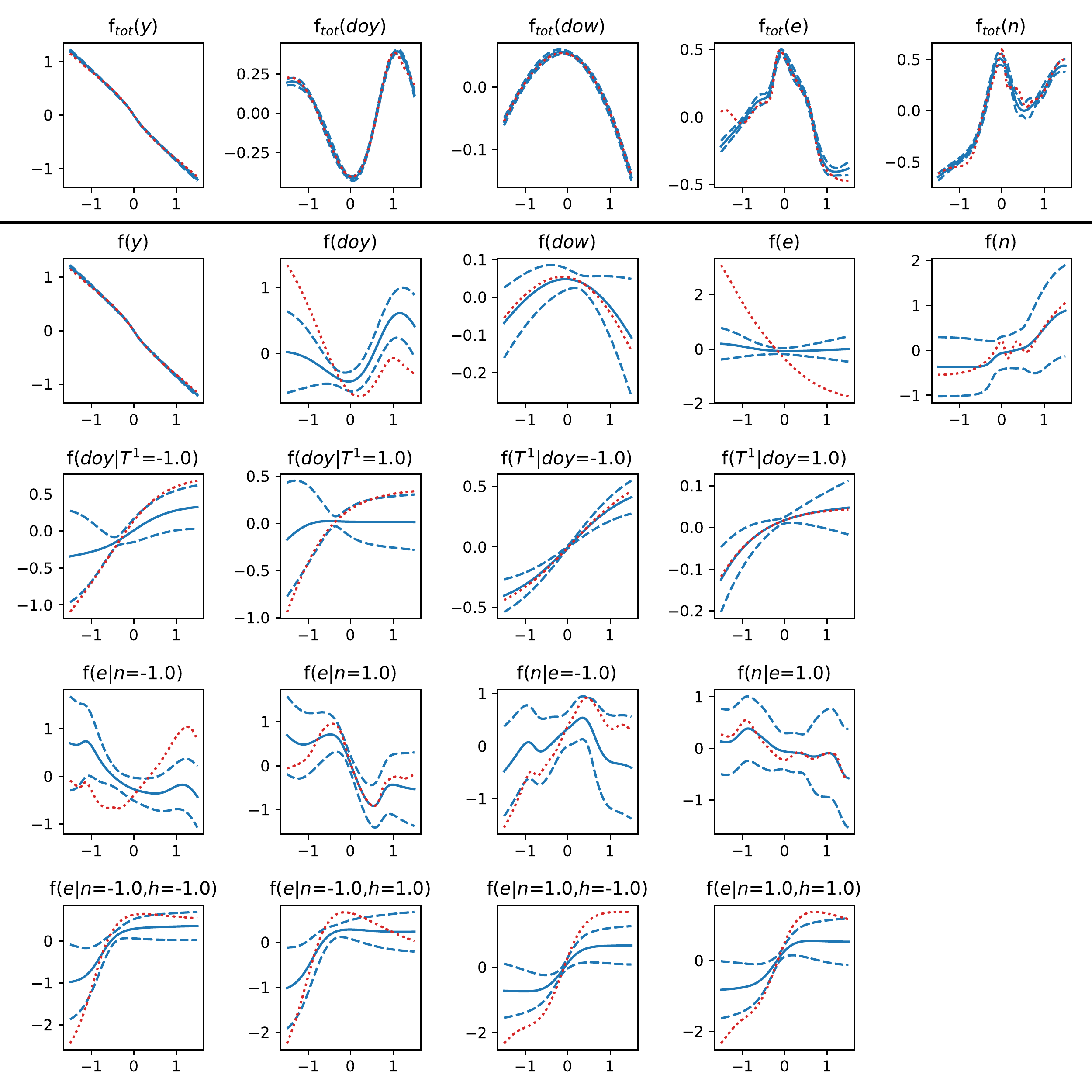}
  \caption{PrAda-net: Comparison of true (red and dotted) and the mean $\pm$ two standard deviations of the identified functions (blue). For the multi-dimensional functions, in each plot all but one of the input variables are kept fixed. The first row shows the total dependency of the covariate, including all functions, while the rest of the rows show the most frequently identified functions.}
  \label{fig:bs_fcts_prada}
\end{figure*}

\begin{figure*}
  \centering
  \includegraphics[width=0.7\textwidth]{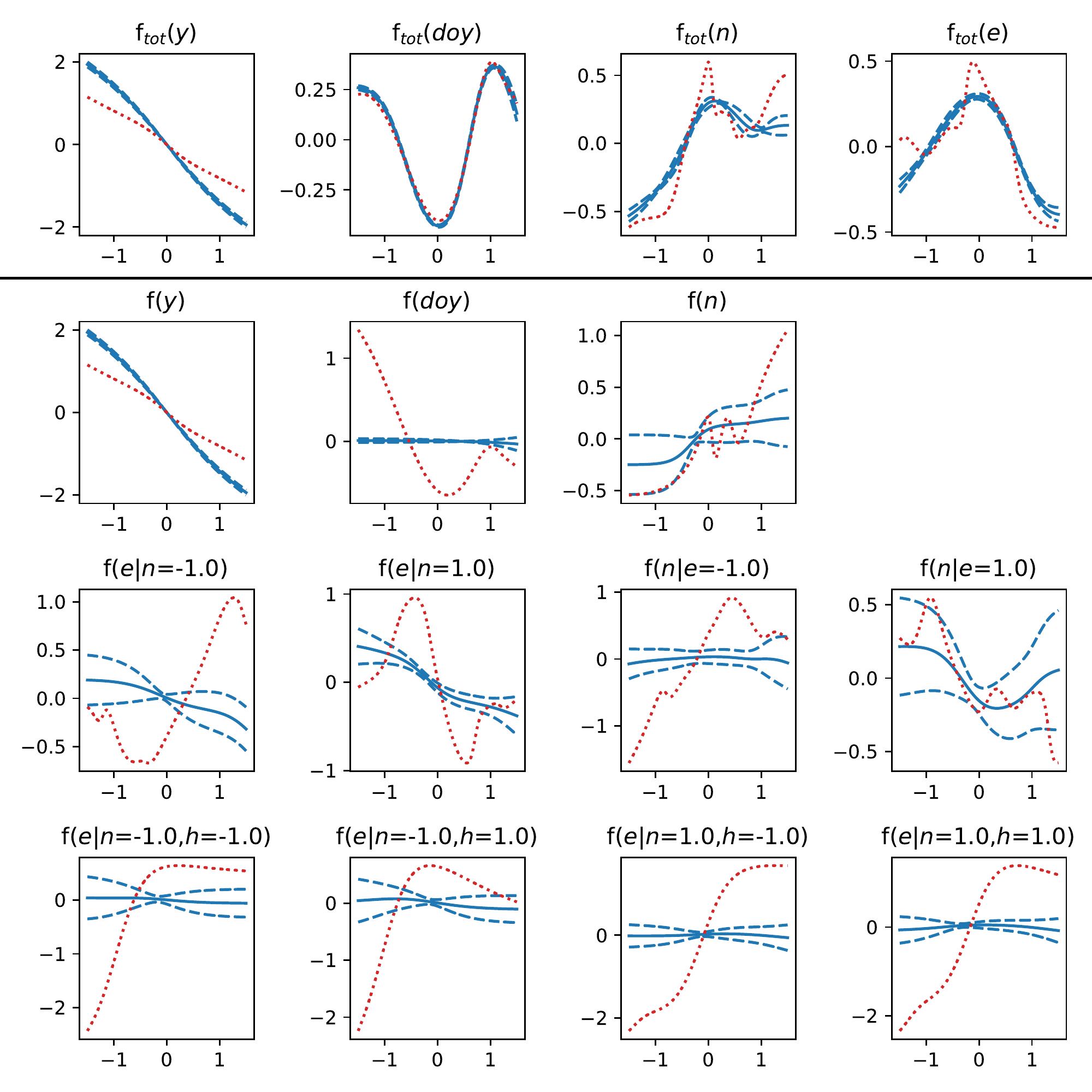}
  \caption{Standard lasso: Comparison of true (red and dotted) and the mean $\pm$ two standard deviations of the identified functions (blue). For the multi-dimensional functions, in each plot all but one of the input variables are kept fixed. The first row shows the total dependency of the covariate, including all functions, while the rest of the rows show the most frequently identified functions.}
  \label{fig:bs_fcts_lasso}
\end{figure*}

In Figures \ref{fig:bs_fcts_prada} and \ref{fig:bs_fcts_lasso}, the means and standard deviations for the most frequently identified functions are plotted together with the true functions. Since the intercept can be realized in any function, all functions were translated to have zero mean. For functions depending on more than one covariate, all covariates except one were fixed at $\pm 1$ to be able to plot in one dimension.
For PrAda-net we see that for total the contribution of a covariate, the true and identified functions coincide very well (first row). This is also true for most identified functions with a few exceptions. However, the differences between true and identified functions seem to be balanced by those of other functions including the same covariate. See e.g.\ $f(\texttt{doy})$ and $f(\texttt{doy}|\texttt{T}^\texttt{1})$, where, for small values of \texttt{doy}, the true $f(\texttt{doy})$ is much larger than the estimated function, while the true $f(\texttt{doy}|\texttt{T}^\texttt{1})$ is much smaller than the estimated function.

Standard lasso exhibits poor overlap with the true model components, even in terms of total contribution. The functions identified by standard lasso align poorly with the true model components.

\section{Conclusions and Discussion}
\label{sec:discussion}
For complex and large data sets, non-parametric additive models often require substantial exploratory work to identify candidate model components, such as sets of potential interactions and format or family of additive function, before any fitting strategy can be applied. Here, we
proposed a fully data-adaptive alternative based on a simple neural network trained with proximal gradient descent and adaptive lasso, PrAda-net.
PrAda-net was found to improve on lasso penalized neural networks both in terms of test error performance and in terms of generating interpretable models. For additive models, PrAda-net was able to identify the function components of the models as well as to express the function complexity by using multiple hidden nodes. 
We illustrated how PrAda-net could be used to model a massive and complex air pollution data set with weather, temporal and spatial covariates. PrAda-net was able to identify a highly similar model to one recently presented in the state-of-the-art literature, but, in contrast, required no pre-selection of modeling components or pre-processing of the data. This data driven strategy thus reduces the dependency on subjective choices and preliminary or partial modeling of complex data, while retaining the interpretability of the classical statistical methods.

While there is no explicit limitation to the depth of the network that PrAda-net can be applied to, we only used one hidden layer in this paper. We thereby loose the possibility of making the network even more compact. However, preliminary results from applying PrAda-net to deep networks indicated that interpretability was much reduced in favor of compact representation. It is not trivial to untangle the compact representation to obtain an interpretable additive model representation. In addition, according to the universal approximation theorem \citep{cybenko1989approximation} the utilization of a one-hidden-layer network does not limit the complexity of the functions that can be modelled. It would, nonetheless, be interesting to apply extensions of PrAda-net to deeper networks, with regularization penalties constructed to e.g.\ guarantee an ordering of main to higher order effects in the layers.  Indeed, structural constraints may be what is needed to generate interpretable networks from more complex architectures. However, this is left for future work.

Code is available at \url{https://github.com/allerbo/prada_net}.

\section*{Acknowledgements}
This research was supported by grants from the Swedish Research Council (VR), the Swedish Foundation for Strategic Research (SSF) and the Chalmers AI Research Center (CHAIR).
\clearpage
\bibliography{prada}
\bibliographystyle{apalike}

\end{document}